# Freezing of Gait Prediction From Accelerometer Data Using a Simple 1D-Convolutional Neural Network

## 8th Place Solution for Kaggle's Parkinson's Freezing of Gait Prediction Competition

Jan Brederecke

Department of Cardiology, University Heart & Vascular Center Hamburg, University Medical Center Hamburg-Eppendorf, Hamburg, Germany
j.brederecke [at] uke.de

**Abstract**

*Freezing of Gait (FOG) is a common motor symptom in patients with Parkinson's disease (PD). During episodes of FOG, patients suddenly lose their ability to stride as intended. Patient-worn accelerometers can capture information on the patient's movement during these episodes and machine learning algorithms can potentially classify this data. The combination therefore holds the potential to detect FOG in real-time. In this work I present a simple 1-D convolutional neural network that was trained to detect FOG events in accelerometer data. Model performance was assessed by measuring the success of the model to discriminate normal movement from FOG episodes and resulted in a mean average precision of 0.356 on the private leaderboard on Kaggle. Ultimately, the model ranked 8th out of 1379 teams in the Parkinson's Freezing of Gait Prediction competition. The results underscore the potential of Deep Learning-based solutions in advancing the field of FOG detection, contributing to improved interventions and management strategies for PD patients.*

## 1. Introduction

The Parkinson's Freezing of Gait Prediction competition was a machine learning competition hosted on Kaggle from March 9, 2023 to June 8, 2023[1].

### *Parkinson's Disease and Freezing of Gait*

Parkinson's disease (PD) is a neurodegenerative disorder characterized among others by motor symptoms, including freezing of gait (FOG), which can severely impact patients' quality of life (Perez-Lloret et al., 2014). FOG is characterized by

*"an episodic inability (lasting seconds) to generate effective stepping in the absence of any known cause other than parkinsonism or high-level gait disorders. It is most commonly experienced during turning and step initiation but also when faced with spatial constraint, stress, and distraction. Focused attention and external stimuli (cues) can overcome the episode"* (Giladi & Nieuwboer, 2008).

With progression of PD, occurrence of FOG in normal walking increases (Falla et al., 2022).

---

[1] https://www.kaggle.com/competitions/tlvmc-parkinsons-freezing-gait-prediction/

The FOG episodes usually last few seconds but can continue for more than 30 s (Schaafsma et al., 2003). FOG leads to mobility impairment and an increased risk of falls in patients with PD, therefore accurate and timely detection of FOG events plays a critical role in providing effective interventions and enhancing the quality of life for individuals with PD.

## The Parkinson's Freezing of Gait Prediction Competition

The objective of the competition was to identify the start and stop of FOG episodes by detecting the occurrence of three types of FOG events: start hesitation (*StartHesitation*), turning (*Turn*), and walking (*Walking*). For this purpose, lower-back 3D accelerometer data from subjects exhibiting FOG episodes was provided.

### Competition Data

Three datasets collected in different settings were available for model training:
- The tDCS FOG (*tdcsfog*) dataset, collected in the lab, as participants completed a FOG-provoking protocol
- The DeFOG (*defog*) dataset, collected in the participant's home, as subjects completed a FOG-provoking protocol
- The Daily Living (*daily*) dataset, collected through one week of continuous 24/7 recordings

The *tdcsfog* and *defog* datasets were annotated by expert reviewers that watched videos of the trials and documented the FOG events. Series in the *daily* dataset were not annotated and it was not used for the development of the presented solution. Each dataset contained three variables related to the acceleration on three axes: V - vertical, ML - mediolateral, AP - anteroposterior. The used sensor data was measured in units of $\frac{m}{s2}$ for *tdcsfog* data and $g$ (9.81 $\frac{m}{s2}$) for *defog* data. Additionally, the *tdcsfog* dataset was recorded at 128 Hz, while the *defog* dataset was recorded using a 100 Hz time resolution. Detailed information on the data and the study population is presented in the competition documentation on Kaggle[2].

### Evaluation

The competition's goal was to detect FOG episodes in withheld test data collected as part of the *tdcsfog* and *defog* settings. No data of patients in the test dataset was provided with the training data to prevent leakage. A solution's performance was measured using the mean average precision score (mAP) across all three outcomes: *StartHesitation*, *Turn*, and *Walking*. The mAP score was calculated as the arithmetic mean of the three average precision scores on the given outcomes. Results are reported on a leaderboard (LB) which is divided into a public LB that provides the score on 32% of the test-data, and the private LB, which provides the score on the remaining hidden test-data (68%). Participants get feedback from the public LB while the private LB score is revealed only after the competition ends. The score on the private LB is what ultimately decides the ranking in the competition.

## 2. Methods

In this section, my solution to the Parkinson's Freezing of Gait Prediction competition is presented. The solution is based on a 1-dimensional convolutional neural network (1D-CNN), that was trained on a combination of the *tdcsfog* and *defog* datasets.

### Preprocessing

Even though the two data sources had different units and slightly different sample rates, data was not transformed to the same unit or normalized. Moreover, no augmentation was used in the training process. For both training and inference, sequences of 1000 timepoints were sampled, 50 of which were located after the events, i.e., in the future.

---

[2] https://www.kaggle.com/competitions/tlvmc-parkinsons-freezing-gait-prediction/overview/data

The *defog* data contained entries with no events, these were excluded for the training. An illustration of the data loader used during training is shown in Figure 1.

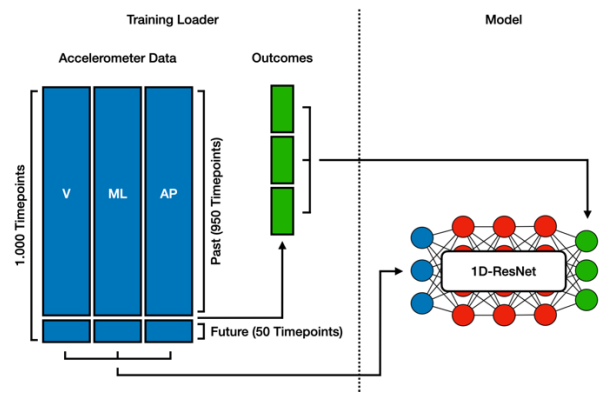

*Figure 1*. Illustration of the Data Loader Used During Training.

## Model Architecture

A ResNet (He et al., 2015) with 1D-CNN layers was used. The model contained six residual block layers with increasing numbers of input / output filters: 32, 64, 128, 256, 384, 512. At the same time the number of input / output samples was decreased: 1000, 500, 250, 125, 25, 5. The model had three input channels, one for each of the variables collected by the sensor. The kernel size was 17. The model was configured to predict the three outcomes *StartHesitation, Turn,* and *Walking*.

## Model Training

Model training was performed in Python 3.8.16[3] using PyTorch 2.0.1 (Paszke et al., 2017). The source code is available under MIT license on the author's GitHub profile[4]. To prevent overfitting, a five-fold cross-validation (CV) scheme was utilized. The CV-scheme used was *Stratified-GroupKFold* from the sci-kit learn (Pedregosa et al., 2011) library which had the additional benefit of preventing the leakage of the same person in the training to the validation splits as the subject was used as the group. For a simplified illustration of the CV see Figure 2.

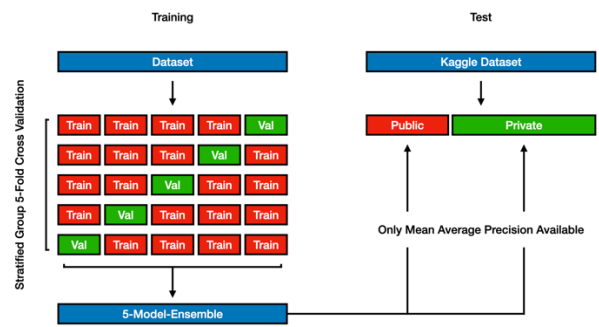

*Figure 2*. Illustration of the Cross-Validation Scheme Used.

Because the task was interpreted as a multi-label classification problem, the loss function used was binary cross entropy for each outcome. PyTorch's *BCEWithLogitsLoss* was therefore utilized due to its known advantages over *BinaryCrossEntropy*[5].

An NVIDIA RTX-3060 GPU with 12 GB VRAM was used for training of the models. All models were trained for a single epoch while limiting the training data to randomly selected five million samples from the respective training folds. A batch-size of 1024, an AdamW optimizer (Loshchilov & Hutter, 2019), and a learning rate scheduler (*ReduceLROnPlateau*[6]) with a starting value of 0.001 were used.

Additionally, Pytorch's integrated ability to use automatic mixed precision[7] was utilized to decrease the overall training time.

---

[3] http://www.python.org/
[4] https://github.com/janbrederecke/fog
[5] https://discuss.pytorch.org/t/bceloss-vs-bcewithlogitsloss/33586/29
[6] https://pytorch.org/docs/stable/generated/torch.optim.lr_scheduler.ReduceLROnPlateau.html
[7] https://pytorch.org/tutorials/recipes/recipes/amp_recipe.html

*Ensembling*

As the CV resulted in five final models, the prediction on the test data was calculated using the simple average of the five individual predictions (see Figure 2).

## 3. Results

The presented solution achieved a mAP score of 0.357 on the public and of 0.356 on the private LB in Kaggle's competition. The local CV resulted in mAP scores ranging from 0.185 to 0.309 (see Table 1). As no other indices could be retrieved from the test data on Kaggle, all other calculations were done using only the local CV scheme. As is shown in Table 1, the individual average precision (AP) for each of the three outcomes in the individual folds ranges from 0.010 to 0.063 for *StartHesitation*, 0.420 to 0.682 for *Turn*, and 0.023 to 0.252 for *Walking*.

*Table 1*. Results of the Cross-Validation.

|  | mAP[2] | AP[3] | | |
| --- | --- | --- | --- | --- |
|  |  | Start Hesitation | Turn | Walking |
| 1[1] | 0.277 | 0.063 | 0.682 | 0.085 |
| 2[1] | 0.261 | 0.010 | 0.562 | 0.212 |
| 3[1] | 0.185 | 0.054 | 0.420 | 0.082 |
| 4[1] | 0.309 | 0.029 | 0.644 | 0.252 |
| 5[1] | 0.203 | 0.013 | 0.575 | 0.023 |

*Notes*. [1]Number of fold. [2]Mean average precision. [3]Average precision.

## 4. Discussion

In order to improve the quality of life of PD patients, reliable automatic FOG detection is an important aspect to consider. The present study reported a highly ranked solution to the Parkinson's Freezing of Gait competition. Results show that a simple 1D-CNN can be successfully used to detect FOG related events in accelerometric data from different settings without applying sophisticated preprocessing.

*Limitations*

While the presented solution achieved a winning place in the competition, it is important to acknowledge that the best-performing models in the competition outperformed my solution significantly. The top-ranking models demonstrated superior mAP (1st place private LB mAP: 0.514), indicating these other approaches may yield better results in overall FOG detection. Especially models that used recurrent neural networks and transformer architectures did well, highlighting their possible superiority over the CNN architecture in FOG prediction from accelerometer data.

Another limitation of the presented solution is the inability to calculate and compare additional performance metrics on the test data. As per the competition guidelines, the test data remain confidential, accessible only to the competition organizers. Consequently, I was unable to calculate common metrics on the individual outcomes such as AUC, ROC curves, or F1 score, which could have provided a more comprehensive evaluation of my model's performance on the test data.

Moreover, while my model performed well in the competition, it is essential to point out the significant differences in AP for the FOG event types. While the model confidently detects *Turn* events across folds, the other events are predicted with comparatively low AP. Future work would need to try to raise individual AP to the level of *Turn*.

In addition to the given limitations, it is important to address that I included 50 consecutive timepoints following the predicted FOG events in my analysis. However, including future timepoints would not be feasible for real-time application, therefore making evaluation of the model on past data only necessary for this purpose.

*Strengths*

Despite the limitations, the reported solution shines through its simplicity. By utilizing a straightforward Deep Learning architecture like the 1D-ResNet, I demonstrate that achieving competitive results in FOG detection can be accomplished without the need for overly complex models, pre-processing or extensive computational resources. This simplicity not only facilitates ease of implementation but also allows other interested researchers to adopt and build upon the reported methodology with relative ease.

*Conclusions*

I present a comparatively simple yet effective event detection and classification pipeline for detection of FOG that took 8th place of 1379 teams at the Parkinson's Freezing of Gait Prediction competition. The fact that such an effective model could be trained despite the few preprocessing steps speaks once again for the capability of the ResNet architecture that has been demonstrated numerous times in the past. Moreover, as the CV score on the training dataset did not differ much from the final competition scores on unseen data, local CV has once again been shown to be an effective tool in preventing overfitting.

**Acknowledgements**

I would like to thank Kaggle and The Michael J. Fox Foundation for hosting the Parkinson's Freezing of Gait Prediction Competition. Moreover, I thank the three research groups that collected the data: The Center for the Study of Movement, Cognition and Mobility, The Neurorehabilitation Research Group at Katholieke Universiteit Leuven in Belgium, and the Mobility and Falls Translational Research Center at the Hinda and Arthur Marcus Institute for Aging, affiliated with Harvard Medical School in Boston. I also thank the participants of the competition for sharing ideas, insights, and code. A special thank you goes out to Mayukh Bhattacharyya who generously provided an incredible public baseline in the discussions section that kickstarted the solution presented in this article. Moreover, I thank Carla Reinbold, Marius Knorr, Jan Bremer, and Tim Brederecke for their diverse support whilst writing this article.